\pdfoutput=1

\documentclass[11pt,table,xcdraw]{article}

\usepackage[preprint]{acl}

\usepackage{times}
\usepackage{latexsym}

\usepackage[T1]{fontenc}

\usepackage[utf8]{inputenc}

\usepackage{microtype}

\usepackage{inconsolata}

\usepackage{graphicx}

\definecolor{lightgray}{gray}{0.9}
\newcommand{\improvedcell}[2]{%
  \ifdim#1pt>#2pt
    \cellcolor{green!15}#1%
  \else
    #1%
  \fi
}
\usepackage{makecell}
\usepackage{multirow}
\usepackage{multicol}
\usepackage{booktabs}
\usepackage{colortbl}
\usepackage{arydshln}
\usepackage{tabularx}
\usepackage{listings}
\usepackage{colortbl}
\usepackage[most, breakable]{tcolorbox}
\usepackage{enumitem}
\usepackage[detect-all]{siunitx}
\usepackage{tabularx}
\usepackage{geometry}
\usepackage{parskip}
\usepackage{pifont}
\usepackage{caption}
\usepackage[normalem]{ulem}
\usepackage{longtable}

\usepackage{multicol}
\usepackage{ulem}

\tcbuselibrary{skins,breakable}

\newcommand{\greencheck}{\textcolor{green!70!black}{\ding{51}}}
\newcommand{\redcross}{\textcolor{red}{\ding{55}}}

\title{LightPAL: Lightweight Passage Retrieval for Open Domain Multi-Document Summarization}
\author{
 \textbf{Masafumi Enomoto\textsuperscript{1}},
 \textbf{Kunihiro Takeoka\textsuperscript{1}},
 \textbf{Kosuke Akimoto\textsuperscript{1}},
 \\
 \textbf{Kiril Gashteovski\textsuperscript{2}},
 \textbf{Masafumi Oyamada\textsuperscript{1}}
\\
 \textsuperscript{1}NEC Data Science Research Laboratories,
 \textsuperscript{2}NEC Laboratories Europe
\\
 \small{
   \textbf{Correspondence:} \href{masafumi-enomoto@nec.com}{masafumi-enomoto@nec.com}
 }
}

\begin{document}
\maketitle
\begin{abstract}
Open-Domain Multi-Document Summarization (ODMDS) is the task of generating summaries from large document collections in response to user queries. This task is crucial for efficiently addressing diverse information needs from users.
Traditional retrieve-then-summarize approaches fall short for open-ended queries in ODMDS tasks. These queries often require broader context than initially retrieved passages provide, making it challenging to retrieve all relevant information in a single search.
While iterative retrieval methods has been explored for multi-hop question answering (MQA), it's impractical for ODMDS due to high latency from repeated LLM inference. 
Accordingly, we propose LightPAL, a lightweight passage retrieval method for ODMDS. LightPAL leverages an LLM to pre-construct a graph representing passage relationships, then employs random walk during retrieval, avoiding iterative LLM inference.
Experiments demonstrate that LightPAL outperforms naive sparse and pre-trained dense retrievers in both retrieval and summarization metrics, while achieving higher efficiency compared to iterative MQA approaches.
\end{abstract}
\section{Introduction}
\label{sec:intro}

\begingroup
    \renewcommand{\thefootnote}{}
    \footnotetext{This paper is under review.}
    \renewcommand{\thefootnote}{\arabic{footnote}}
\endgroup

\begin{figure}[!t]
    \centering
    \includegraphics[width=7.4cm]{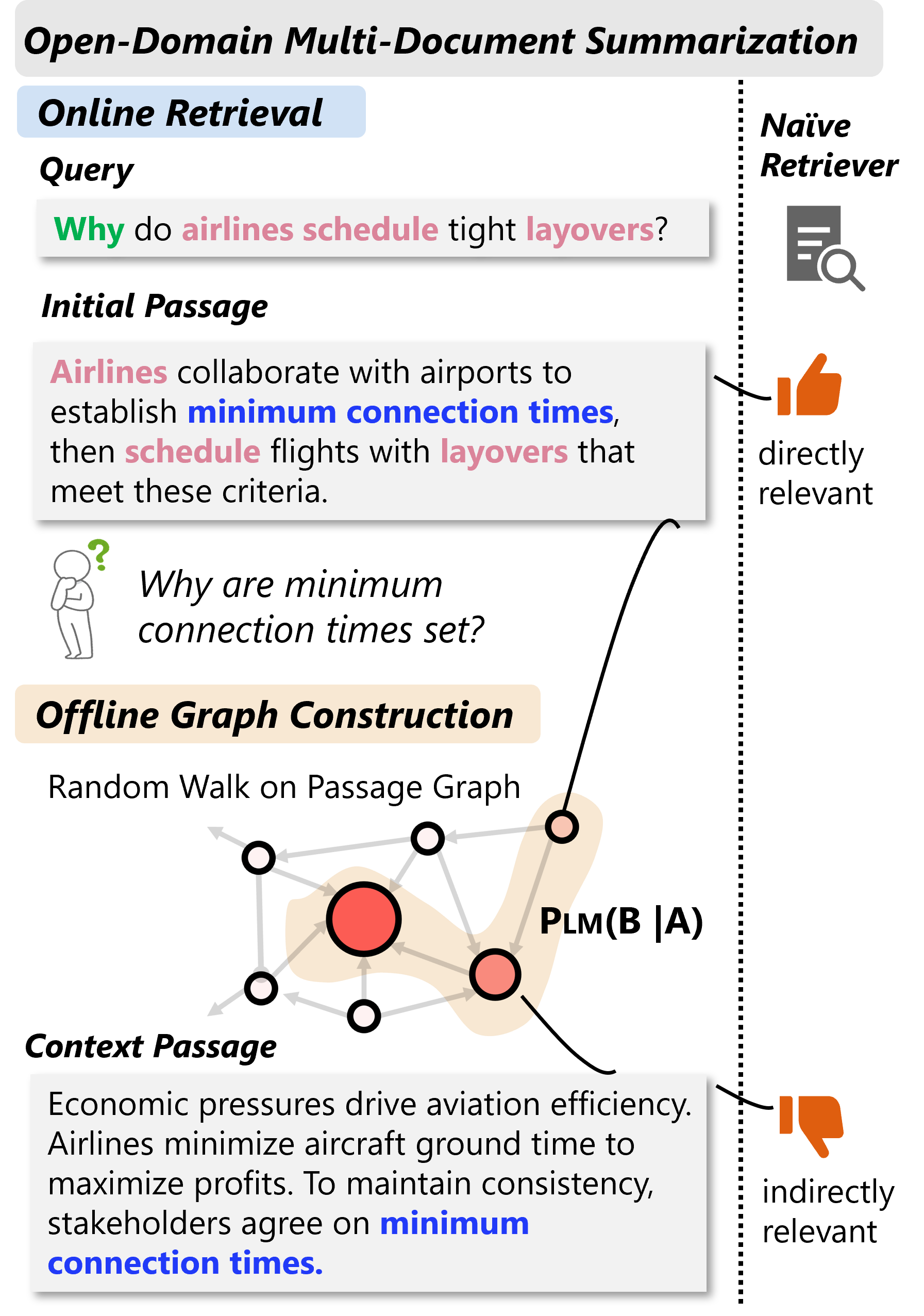}
    \caption{
    LightPAL: lightweight retrieval method for Open-domain Multi-Document Summarization (ODMDS). It consists of two main processes: 1) Offline Graph Construction: evaluates passage relevance using LLM conditional generation probabilities to create edges between passages. 2) Online Retrieval: uses an off-the-shelf retriever for initial passages, then performs a Random Walk on the constructed graph to retrieve informative context passages referenced by many others. This approach efficiently retrieves context passages without iterative LLM inference in runtime.
    }
    \label{fig:problem_statement}
\end{figure}

Summarization is a task that aims to generate text that preserves the most informative content from input documents. As an advanced version, \textbf{Open-Domain Multi-Document Summarization} (ODMDS)~\cite{related/odmds/pioneer, related/odmds/bench} has recently been introduced, where the inputs are a large collection of documents, and the summary should be an answer to a query that reflects the information needs of the user.
ODMDS requires: 
(1) generating a summary in response to queries that ask for an open-ended questions, i.e., queries that demand a summary as an answer.
(2) synthesizing the content of multiple documents related to the query within the large document collection.
Users typically have diverse information needs, making it impractical to precompile a document collection that caters to all possible queries. 
Consequently, ODMDS task plays a vital role in information-seeking scenarios where the system must adapt to information need of each user.

A practical approach to ODMDS is retrieve-then-summarize~\cite{related/odmds/pioneer}, where a retrieval model first finds relevant passages for the query, and then a language model generates a summary. The quality of the summary heavily depends on retrieving comprehensive and relevant information. Existing work~\cite{related/odmds/bench} has employed sparse and pre-trained dense retrievers, similar to open-domain question answering~\cite{related/qa/survey}. However, these retrievers are often insufficient for ODMDS. This is because users seek summaries to efficiently acquire knowledge about unfamiliar topics, with information needs not typically expressed as specific queries like factoid. For instance, consider the question "Why do airlines schedule tight layovers?" (see Figure~\ref{fig:problem_statement}). Users might only realize they need to ask about "minimum connection times" after reading initial results. For such open-ended questions, recursive search is required, where initially retrieved information is used for reasoning to search for further information.

Iterative retrieval and reasoning methods have been primarily proposed for multi-hop question answering (MQA), with recent methods leveraging the reasoning capability of large language models (LLMs)~\cite{related/mqa/TrivediBKS23, related/mqa/KhalifaLLL023}. 
We applied these approaches to ODMDS, demonstrating improved retrieval performance and summary quality (Section~\ref{sec:exp}). However, ODMDS presents unique challenges due to the larger number of relevant passages required for open-ended questions compared to MQA.
For example, the Story dataset for ODMDS~\cite{related/odmds/bench} averages 9 relevant passages per query, whereas HotPotQA~\cite{related/mqa/hotpotqa} for MQA typically requires only two. 
This necessitates numerous iterations in ODMDS, leading to high latency when using LLM inference at each iteration.
Consequently, a more efficient method is needed to balance summary quality and computational cost in the ODMDS context.

In this paper, we propose \textbf{LightPAL}, a lightweight passage retrieval method for ODMDS. It constructs a graph representing passage relationships using a LLM during indexing, then employs random walk via Personalized PageRank (PPR)~\cite{method/orig_ppr} for retrieval. During indexing, the LLM evaluates if a passage could serve as context for another, creating links based on conditional generation probabilities. This captures relevance between passages by considering diverse styles and domains. At retrieval, an initial set of relevant passages is obtained using an arbitrary model. PPR then calculates the probability of reaching other passages from the initial set on the graph. This enables low-latency retrieval of diverse, relevant information without LLM inference during runtime.

We evaluated LightPAL using two benchmarks: ODSum~\cite{related/odmds/bench} for ODMDS and LFRQA~\cite{experiment/dataset/LFRQA} for long-form question answering in retrieval-augmented generation (RAG). Experiment results showed that LightPAL improved retrieval performance by up to 10\% compared to sparse/pre-trained dense retrievers. Results also indicated that LightPAL performed comparably to or outperformed PromptRank~\cite{related/mqa/KhalifaLLL023}, an unsupervised iterative reranking method for MQA, while achieving about 1,000 times lower search latency~\footnote{\texttt{Qwen2.5-3B} is used as an LLM in PromptRank.}. For generated summary quality, LightPAL outperformed baseline retrievers when retrieval performance was substantially enhanced, as measured by ROUGE metrics and GPT-4-based evaluation~\cite{experiment/metric/geval}. We conducted human annotations on a subset of dataset to validate the GPT-4 evaluation, which revealed a moderate alignment between GPT-4 and human assessments.
\section{Related Works}
\label{sec:related}

\noindent
\textbf{Open-domain Multi-document Summarization. }
Open-Domain Multi-Document Summarization (\textbf{ODMDS})~\cite{related/odmds/pioneer, related/odmds/bench} is a task that summarizing information from large collection of documents into a summary as answer to given a query. \citet{related/odmds/pioneer} introduced ODMDS and created a pseudo-ODMDS dataset using existing multi-document summarization data to evaluate the capabilities of a retrieve-then-summarize pipeline. The results revealed a decline in summary quality due to the retrieval phase. More recently, \citet{related/odmds/bench} created a benchmark for the ODMDS task and evaluated the performance of LLMs as summarization models. 
These initial studies have examined basic retrieval and summarization components of RAG systems for ODMDS. However, no research has addressed the development of iterative retrieval methods for open-ended queries in ODMDS or their efficiency.

\noindent
\textbf{Retrieval for Multi-hop/document QA. }
Multi-hop Question Answering (\textbf{MQA}) tasks~\cite{related/mqa/survey} require integrating information from multiple passages to obtain an answer.
Existing approaches typically employ iterative reasoning and retrieval for multi-hop searches.
\citet{related/mqa/DasDZM19} update query vectors using the reader model's state for reranking passages, while \citet{related/mqa/QiLMWM19} extract text spans from retrieved passages to query additional passages.
Recent methods leverage LLMs to assess relevance between queries and passages. \citet{related/mqa/KhalifaLLL023} proposed an unsupervised re-ranking method, scoring hyperlinked passages based on LLMs' query generation probability.
As a more advanced method, researchers use LLMs as agents to traverse pre-constructed graphs of passages.
Knowledge Graph Prompting~\cite{related/rag/graph_trav/KGP} uses LLMs to generate query-answering information and select similar neighboring passages.
GraphReader~\cite{related/rag/graph_trav/GraphReader} extends this for exploration planning, relevance evaluation, and content selection.
Despite effectiveness, these methods face real-world limitations due to high latency from iterative LLM inference.

\noindent
\textbf{Indexing for Comprehensive Question Answering in RAG Framework. }
Recent RAG frameworks organize and condense documents prior to retrieval for comprehensive question answering. RAPTOR~\cite{related/rag/graph_ret/raptor} constructs tree structures through recursive summarization, while GraphRAG~\cite{related/rag/graph_ret/GraphRAG} creates entity graphs for hierarchical dataset summaries. These methods, like our approach, improve indexing to use contextual information in retrieval. However, they are unsuitable for our aim as they do not address the ODMDS task or target latency reduction in retrieval. GraphRAG is unsuitable due to its extensive use of LLMs for relevance scoring against multiple summaries, leading to high latency. RAPTOR, designed for single-document QA, does not apply to ODMDS scenarios where passages are from various documents.
\section{ODMDS Task}
Open-Domain Multi-Document Summarization (ODMDS) is formally defined as follows: given a natural language query $q$ and a large collection of passages $D=\{d_i\}_i$, generate a summary $S$ corresponding to the query $q$.
Following prior work~\cite{related/odmds/bench}, we adopt retrieve-then-summarize approach for the ODMDS task~\footnote{While some LLMs, such as Gemini Pro 1.5(\url{https://deepmind.google/technologies/gemini/pro/}), can handle up to 1M input tokens, the size of the passage collection is generally unbounded, making retrieval a necessity.}. First, we retrieve a set of passages $D_q \subset D$ relevant to the query $q$ from the collection $D$. Then, we input the retrieved text and query to a language model $LM_{sum}$ to generate an abstractive summary $S = LM_{sum}(q, D_q)$.
\section{LightPAL: Proposed Method}
\label{sec:method}
We propose a retrieval method for addressing open-ended ODMDS queries. Our approach retrieves relevant and diverse passages while avoiding iterative retrieval and language model inference at runtime. 
As shown in Figure~\ref{fig:problem_statement}, The method consists of two phases:
\textbf{(1) Graph Construction}: Pre-constructing a graph representing passage relationships using a large language model (LLM), linking passages with high conditional generation probabilities as potential context for a passage. 
\textbf{(2) Retrieval}: Retrieving an initial set of relevant passages with a conventional retriever, then employing random walk via Personalized PageRank (PPR)~\cite{method/orig_ppr} on the pre-built graph. PPR calculates probabilities of reaching other passages from initial passages, retrieving highly probable ones as additional context without runtime language model inference.

\noindent
\textbf{Passage Graph Construction. }
The motivation behind our approach is that passages extracted from a larger document may not be self-explanatory and require context from other passages to be correctly understood. Even cohesive passages can benefit from related texts to deepen topic understanding. Naive retrieval systems that only consider query-passage relevance cannot capture these relationships. We aim to address this by creating a graph annotated with passage relationships using the conditional generation probabilities of LLMs.
Concretely, for each ordered pair of passages $(d_{i}, d_{j})$ in passage set $D$, we calculate the conditional generation probability of passage $d_{j}$ given passage $d_{i}$ as a context score using a language model: $$\text{ContextScore}(d_{i}, d_{j}) = P_{LM}(d_{j}|d_{i}).$$
We construct a passage graph $G$ by creating edges between the passage $d_i$ and the top-$k$ passages $\{ d_j \}_j$ with the highest $\text{ContextScore}(d_i, d_j)$.
Calculating context scores for all passage pairs is computationally prohibitive. 
Thus, we first retrieve the top-$k^\prime$ candidate passages similar to each passage using a lightweight embedding model and then run the language model on these candidates to create edges for the top-$k$ ($k\ll k^\prime$) passages with the highest context scores.

We choose the conditional generation probabilities of LLMs to build a passage graph because they can richly determine the relevance between passages by considering the diverse styles and domains of the passages in the passage set $D$. 
They capture the fluency of the passage sequence, making them suitable for reconstructing relationships between passages originally part of a cohesive document or for building new relationships between passages from different documents that contribute as the context.

\noindent
\textbf{Passage Retrieval. }
Our proposed method consists of two phases for passage retrieval: 1) \textbf{Initial Passage Retrieval}: Retrieve passages $D_{init}$ directly relevant to the query using any sparse/dense retrieval model. 2) \textbf{Context Passage Retrieval}: Using the initial passages, additionally retrieve context passages $D_{context}$ from the passage graph $G$ that provide contextual information to the initial passages. Finally, we obtain $D_q$ as the union of $D_{init}$ and $D_{context}$.
For context passage retrieval, we employ Personalized PageRank (PPR)~\cite{related/misc/PPR}. PPR calculates the probability of reaching other passages from the initial set $D_{init}$ via a random walk on the graph $G$, highly ranking passages cohesively referenced by many passages. Starting a random walk from the query-relevant initial passages, we expect to explore topics related to the query, and then retrieve highly informative context passages for those topics using PPR. This approach enables low-latency retrieval of relevant information without iterative search using LLM inference during runtime.
\begin{table}[!t]
    \centering
    \begin{tabular}{lrrr}
        \toprule
        \textbf{} & \textbf{Story} & \textbf{Meeting} & \textbf{RAG-QA} \\
        \midrule
        \# Queries & 260 & 131 & 234 \\
        \# Passages & 1190 & 13074 & 3358 \\
        \# PassLen. & 3171 & 899 & 834 \\
        \# SummLen. & 1286 & 1081 & 1503 \\
        \# Rel. & 9.2 & 177.8 & 10.1 \\
        Rel. \% & 0.8\% & 1.4\% & 0.3\% \\
        \bottomrule
    \end{tabular}
    \caption{Statistics of the datasets. Abbreviations used: PassLen. (Average Passage Length in characters), SummLen. (Average Reference Summary Length in characters), Rel. (Average Relevant Passages), Rel. \% (Relevant Passages Ratio).}
    \label{tab:odsum_datasets}
\end{table}

\section{Experiments}
\label{sec:exp}
\subsection{Settings}
\label{sec:exp/setting}
\textbf{Data.} 
To evaluate our approach, we used three datasets:
(1) \textbf{Story} dataset from ODSum~\cite{related/odmds/bench}, where passages are sections from independent stories, requiring retrieval of cohesive story sections relevant to the query.
(2) \textbf{Meeting} dataset from ODSum, where queries are questions about meeting contents. We divided each long meeting transcript (approx. 8k tokens) into smaller chunks of around 1k characters, treating each chunk as a passage to avoid irrelevant information degrading summarization performance. If meeting transcript is labeled as relevant to query, chunks are also labeled as relevant. We used test set from both Story and Meeting.
(3) \textbf{RAG-QA} dataset, derived from LFRQA~\cite{experiment/dataset/LFRQA}, a benchmark for evaluating retrieval-augmented generation systems in long-form question answering. We linked LFRQA data with the LoTTE~\cite{experiment/dataset/LoTTE} dataset to obtain relevant passage sets for each query.  Queries are StackExchange question titles and Google search-autocomplete queries, while passages are StackExchange answers. We sampled up to 50 queries from five domains (lifestyle, recreation, science, technology, and writing), each with at least five related documents and answer lengths over 1k characters. We also added randomly sampled negative passages to create a comprehensive document collection. This dataset evaluates LightPAL's performance in diverse, realistic information-seeking scenarios.

\noindent
\textbf{Baselines for Passage Retrieval.}
To investigate the impact of retrieval methods on the final summary quality, we prepared two types of retrieval methods as naive baselines: a pre-trained dense retriever and a sparse retriever. The dense retriever embeds passages and queries into the same vector space and retrieves the top-K passages most similar to the query. We used the \textbf{bge-large-en-v1.5}~\footnote{\url{https://huggingface.co/BAAI/bge-large-en-v1.5}} model for embedding as it is a top-tier model in Massive Text Embedding Benchmark~\footnote{\url{https://github.com/embeddings-benchmark/mteb}}. For the sparse retriever, we employed \textbf{BM25} as an alternative baseline.

For Context Passage Retrieval, we compared LightPAL with \textbf{PromptRank}~\cite{related/mqa/KhalifaLLL023}, an unsupervised reranker for Multi-hop QA. PromptRank performs iterative reasoning and retrieval by traversing links between passages. It requires links between passages, so to enable a fair comparison, we employ the passage graph created by our method. 
For a fair comparison, both PromptRank and our proposed method initially used the top $|D_{init}|$ passages retrieved by either the dense retriever or BM25. Each method then retrieved the remaining $|D_{context}| = K - |D_{init}|$ context passages using their respective approaches. We set $|D_{init}|$ as top 60\% of $K$.

\noindent
\textbf{Detailed Settings of LightPAL. }
For graph construction of LightPAL, we used the \texttt{bge-large-en-v1.5} embedding model to obtain the top-100 most similar passages for each passage as candidates. For a fair comparison with PromptRank, we calculated the $\text{ContextScore}$ for these passage pairs using the same language models employed by PromptRank.
We then constructed a graph by creating edges between the top-5 pairs based on their ContextScore.
In the Personalized PageRank algorithm, edge weights were uniformly set, and the algorithm was configured to jump uniformly to the top-20 of initial passages $D_{init}$ with a probability of $1 - \alpha$, where $\alpha$ was set to 0.2.

\noindent
\textbf{Number of Passages and Language Models for Summary Generation.} 
For the ODSUM task, we speculate that the number of retrieved passages affects the final summary quality. Thus, we tested three scenarios: the retrieved passages are fewer than, approximately equal to, or more than the average number of relevant passages in each dataset (see Table~\ref{tab:odsum_datasets} for detailed statistics). In this way, we experimented with multiple retrieval settings: 5, 10, and 20 passages for Story and RAGQA data, and 100, 200, and 300 ones for Meeting data. Settings with higher numbers require processing larger volumes of text. To handle this long-context setting, we employed top-tier long-context LLMs from the RULER benchmark~\cite{related/misc/RULER} as our summary generators: \texttt{Meta-Llama-3.1-8B-Instruct}\footnote{\url{https://huggingface.co/meta-llama/Llama-3.1-8B-Instruct}} and \texttt{MegaBeam-Mistral-7B-512k}\footnote{\url{https://huggingface.co/aws-prototyping/MegaBeam-Mistral-7B-512k}}. The prompts used for summary generation are provided in Appendix~\ref{appendix:prompts/summary}.

\begin{table*}[!t]
\centering
\begin{tabular}{l|cc|cc|cc|cc}
\toprule
\multirow{2}{*}{Method} & \multicolumn{2}{c|}{Story} & \multicolumn{2}{c|}{Meeting} & \multicolumn{2}{c}{RAG-QA} \\
 & Pre@k & Rec@k & Pre@k & Rec@k & Pre@k & Rec@k \\
\midrule
\textbf{bge-large-en-v1.5} & 31.15 & 36.27 & 10.34 & 12.79 & \textbf{\underline{64.25}} & \textbf{\underline{71.49}} \\
+PromptRank (Qwen 0.5B) & 31.51 & 37.36 & \textbf{\underline{12.22}} & 11.16 & 59.77 & 67.71 \\
+PromptRank (Qwen 1.5B) & 34.96 & 40.77 & 10.94 & 12.70 & 59.99 & 68.06 \\
+PromptRank (Qwen 3B) & 36.37 & 42.24 & \textbf{11.48} & 12.52 & 59.89 & 68.24 \\
\rowcolor{lightgray}+LightPAL (Qwen 0.5B) & 37.16 & 42.73 & 11.37 & 14.26 & 62.78 & 70.58 \\
\rowcolor{lightgray}+LightPAL (Qwen 1.5B) & \textbf{38.50} & \textbf{44.24} & 11.41 & \textbf{14.32} & 63.14 & 70.73 \\
\rowcolor{lightgray}+LightPAL (Qwen 3B) & \textbf{\underline{39.55}} & \textbf{\underline{45.30}} & 11.47 & \textbf{\underline{14.38}} & \textbf{63.43} & \textbf{70.84} \\
\midrule
\textbf{BM25} & 43.85 & 49.76 & 11.04 & 13.36 & 50.23 & 54.91 \\
+PromptRank (Qwen 0.5B) & 40.04 & 46.92 & 12.27 & 12.85 & 53.34 & 59.67 \\
+PromptRank (Qwen 1.5B) & 42.63 & 49.31 & 11.66 & 13.43 & 54.00 & 60.63 \\
+PromptRank (Qwen 3B) & 43.25 & 50.12 & \textbf{\underline{12.44}} & 13.54 & 53.75 & 60.37 \\
\rowcolor{lightgray}+LightPAL (Qwen 0.5B) & 48.75 & 55.05 & 12.30 & \textbf{15.17} & 54.58 & 60.90 \\
\rowcolor{lightgray}+LightPAL (Qwen 1.5B) & \textbf{49.72} & \textbf{55.94} & 12.18 & 15.13 & \textbf{54.95} & \textbf{61.42} \\
\rowcolor{lightgray}+LightPAL (Qwen 3B) & \textbf{\underline{50.22}} & \textbf{\underline{56.47}} & \textbf{12.36} & \textbf{\underline{15.30}} & \textbf{\underline{55.66}} & \textbf{\underline{61.88}} \\
\bottomrule
\end{tabular}
\caption{
Retrieval performance comparison among PromptRank and LightPAL with dense (bge-large-en-v1.5) and sparse (BM25) retrievers. 
Metrics include Precision@k and Recall@k (abbreviated as Pre@k and Rec@k in the table). Within each group, the highest score is shown in bold and underlined; the second-highest score is in bold. Scores are averaged across different numbers of retrieved passages.}
\label{tab:performance_retrieval}
\end{table*}

\subsection{Retrieval Performance Evaluation}
\label{sec:exp/retrieval}
\noindent
\textbf{Evaluation Metric.}
To first confirm that our proposed method enhances retrieval performance, we evaluated Precision@$k$ and Recall@$k$ for the top-$k$ retrieved passages.
Precision@$k$ measures the fraction of passages among the top-$k$ retrieved results relevant to the query, and Recall@$k$ does the fraction of all relevant passages successfully retrieved within the top-$k$ results.

\noindent
\textbf{Effectiveness.}
Both LightPAL and PromptRank utilize LLMs, and we hypothesized that the retrieval performance might vary depending on the capacity of these models. To investigate this, we conducted experiments using models of different sizes from the same series: \texttt{Qwen2.5-0.5B}, \texttt{Qwen2.5-1.5B}, and \texttt{Qwen2.5-3B}\footnote{\url{https://github.com/QwenLM/Qwen2.5}}.
Table~\ref{tab:performance_retrieval} shows LightPAL's retrieval performance using dense or sparse retrievers as base models. LightPAL improves performance for most dataset and base model combinations, except for RAG-QA and bge-large-en-v1.5. Performance enhancement is most significant for Story data, improving up to 10\%. LightPAL performs comparably or better than PromptRank, outperforming it by 3-6\% on Story/RAG-QA data. 
The experimental results also demonstrated the impact of LLM size on retrieval performance in LightPAL and PromptRank: using larger LLM models gradually improved retrieval performance, especially for Recall@$k$ metrics. However, performance gains from larger models are relatively small compared to improvements over baseline models. This suggests that even models smaller than 0.5B may still offer substantial benefits.

\noindent
\textbf{Search Latency.}
Table~\ref{tab:efficiency_search} presents the average latency for retrieving initial passages and context passages for a single query. The results demonstrate that LightPAL is approximately 1,000 times faster than PromptRank. Moreover, PromptRank's latency exceeds 10 seconds, which is prohibitively slow for interactive RAG systems. In contrast, LightPAL's latency is only about 10 to 20 times higher than the initial search's. This suggests that LightPAL is unlikely to impair the real-time performance of RAG systems. The time complexity for computing Personalized PageRank scores is linear in the number of edges in the graph~\cite{related/misc/PPR_algo}, which allows for practical search times.

\noindent
\textbf{Graph Construction Time.}
Table \ref{tab:efficiency_graphconst} shows the time required for graph construction in LightPAL. These results demonstrate that graph construction can be achieved through lightweight processing. For instance, using Qwen2.5-0.5B on the Meeting dataset (approximately 10k passages), the process completes in about 4 hours on 8 NVIDIA L40S GPUs. Our results show that LightPAL improves retrieval performance while maintaining low search latency through offline graph construction.

\begin{table}[h]
\centering
\begin{tabular}{l|cc}
\hline
& \multicolumn{2}{c}{\textbf{Init. PR [ms]}} \\
data & bge-large-en-v1.5 & BM25 \\
\hline
Story & 2.23 & 1.59 \\
Meeting & 5.24 & 32.31 \\
RAG-QA & 2.50 & 2.89 \\
\midrule
& \multicolumn{2}{c}{\textbf{Cont. PR [ms]}} \\
& PromptRank & LightPAL \\
\hline
Story & $7.57 \times 10^{4}$ & 6.02 \\
Meeting & $3.09 \times 10^{4}$ & 113.31 \\
RAG-QA & $1.96 \times 10^{4}$ & 23.09 \\
\hline
\end{tabular}
\caption{Average search latency for Initial Passage Retrieval (Init. PR) and Context Passage Retrieval (Cont. PR). For Init. PR, latency includes both embedding and search time. Query embedding for bge-large-en-v1.5 was performed on NVIDIA L40 GPU. Search operations for bge-large-en-v1.5, BM25, and LightPAL were conducted on a single thread of AMD EPYC 9554 processor. For PromptRank, LLM inference (using Qwen2.5-3B model) was executed on a single NVIDIA L40S GPU.}
\label{tab:efficiency_search}
\end{table}

\begin{table}[h]
    \centering
    \resizebox{\columnwidth}{!}{
        \begin{tabular}{lrrr}
            \toprule
            \textbf{} & \textbf{Story} & \textbf{Meeting} & \textbf{RAG-QA} \\
            \midrule
            Qwen2.5-0.5B & 0h 22m & 3h 43m & 0h 57m \\
            Qwen2.5-1.5B & 0h 44m & 7h 40m & 1h 58m \\
            Qwen2.5-3B & 1h 11m & 12h 30m & 3h 13m \\
            \bottomrule
        \end{tabular}
    }
    \caption{Computational time to calculate ContextScore for each language model in LightPAL. The preceding and following passages were truncated to a combined length of 1024 tokens. Inference was performed using data parallelism across 8 NVIDIA L40S GPUs.}
    \label{tab:efficiency_graphconst}
\end{table}

\begin{table}[t]
\centering
\newcommand{\lightblue}{\cellcolor{blue!15}}
\newcommand{\lightred}{\cellcolor{red!15}}
\resizebox{\columnwidth}{!}{
    \begin{tabular}{l| l l l c}
    \toprule
    Method & Rel & Comp & Div & ROUGE \\
    \midrule
    \midrule
    \textbf{Story} &&&&\\
    \cellcolor{gray!15}\textit{bge-large-en-v1.5} & 2.34 & 3.46 & 2.14 & 10.86 \\
    +PromptRank & \lightblue2.47 & \lightblue3.47 & \lightred2.11 & \lightblue11.20 \\
    +LightPAL & \lightblue2.44 & \lightblue3.50 & \lightblue2.16 & \lightblue10.99 \\
    \midrule
    \cellcolor{gray!15}\textit{bm25} & 2.56 & 3.58 & 2.23 & 11.54 \\
    +PromptRank & \lightblue2.59 & \lightred3.57 & \lightred2.17 & \lightblue11.56 \\
    +LightPAL & \lightblue2.59 & \lightblue3.60 & \lightred2.22 & \lightred11.53 \\
    \midrule
    \midrule
    \textbf{Meeting} &&&&\\
    \cellcolor{gray!15}\textit{bge-large-en-v1.5} & 1.79 & 3.33 & 2.69 & 11.95 \\
    +PromptRank & \lightblue1.82 & \lightblue3.36 & \lightblue2.73 & \lightblue12.21 \\
    +LightPAL & \lightred1.78 & \lightblue3.34 & \lightblue2.71 & \lightred11.87 \\
    \midrule
    \cellcolor{gray!15}\textit{bm25} & 1.79 & 3.20 & 2.64 & 11.93 \\
    +PromptRank & \lightblue1.81 & \lightblue3.29 & \lightblue2.66 & \lightblue12.30 \\
    +LightPAL & \lightred1.77 & \lightblue3.23 & \lightblue2.66 & \lightblue11.95 \\
    \midrule
    \midrule
    \textbf{RAG-QA} &&&&\\
    \cellcolor{gray!15}\textit{bge-large-en-v1.5} & 2.70 & 3.81 & 3.38 & 19.16 \\
    +PromptRank & \lightred2.63 & \lightred3.75 & \lightred3.34 & \lightred19.03 \\
    +LightPAL & \lightblue2.71 & \lightred3.80 & \lightred3.36 & \lightblue19.17 \\
    \midrule
    \cellcolor{gray!15}\textit{bm25} & 2.55 & 3.62 & 3.10 & 18.06 \\
    +PromptRank & \lightblue2.63 & \lightblue3.70 & \lightblue3.21 & \lightblue18.72 \\
    +LightPAL & \lightblue2.64 & \lightblue3.66 & \lightblue3.19 & \lightblue18.56 \\
    \bottomrule
    \end{tabular}
}
\caption{Comparison of summary quality across different datasets and base retrievers. Scores that improved from the base retriever are displayed in \textcolor{blue}{blue}, while others are in \textcolor{red}{red}. Summary quality metrics (Relevance, Comprehensiveness, and Diversity) are evaluated using G-Eval~\cite{experiment/metric/geval}, which assigns weights to scores 1-5 based on GPT-4's generation probabilities. The displayed scores are averaged across various numbers of retrieved passages and language models used for summary generation.}
\label{tab:performance_summarization_auto}
\end{table}

\begin{table}[t]
\centering
\begin{tabular}{l|cc|cc|cc}
\toprule
\multirow{2}{*}{Dataset} & \multicolumn{2}{c|}{Rel} & \multicolumn{2}{c|}{Comp} & \multicolumn{2}{c}{Div} \\
& $\rho$ & $\tau$ & $\rho$ & $\tau$ & $\rho$ & $\tau$ \\
\midrule
Story    & .47 & .36 & .49 & .40 & .50 & .38  \\
RAG-QA   & .36 & .27 & .48 & .39 & .18 & .13  \\
\bottomrule
\end{tabular}
\caption{Summary-level correlation coefficients (Spearman's $\rho$ and Kendall's $\tau$) between human judgments and GPT-4 judgments for different datasets and aspects.}
\label{tab:correlation}
\end{table}

\subsection{Summary Quality Comparison}
\label{sec:exp/summ_quality}
\noindent
\textbf{Evaluation Metric.}
To evaluate the generated summaries, we calculated the average of ROUGE-1, -2, and -L scores for relevance to reference summaries. We also used G-Eval~\cite{experiment/metric/geval}, which weights scores 1-5 based on GPT-4's generation probabilities, to assess three aspects:
\textbf{(1) Relevance}: How much important content from the reference summaries is covered. Assessed whether the summary captures the key points and essential information from the reference summaries.
\textbf{(2) Diversity}: How varied and insightful the summary is in providing different perspectives on the question. Assessed whether the summary introduces multiple viewpoints and facilitates a broader understanding of the topic.
\textbf{(3) Comprehensiveness}: How well the summary covers all aspects of the question. Evaluated whether the summary provides a thorough representation of information that directly addresses the question.
The prompts used for evaluation are provided in Appendix~\ref{appendix:prompts/geval}.

\noindent
\textbf{Results.}
The comparison results of Summary Quality are presented in Table~\ref{tab:performance_summarization_auto}. These results demonstrate that LightPAL improves summarization performance in cases where retrieval performance was notably enhanced. While slightly inferior to PromptRank, LightPAL still shows improvements in overall summary quality. For the \textbf{Story} dataset, LightPAL achieves noticeable enhancements, particularly with bge-large-en-v1.5 base retriever. Both Relevance and ROUGE scores increase by approximately 0.1 points, aligning with the substantial improvement in retrieval performance. In contrast, the \textbf{Meeting} dataset shows only marginal improvements with LightPAL, consistent with minimal retrieval performance enhancement. For the \textbf{RAG-QA} dataset, using bm25 as the base retriever improves summarization performance: Relevance and Diversity increase by about 0.1 points, and ROUGE by 0.5 points. However, with the bge-large-en-v1.5 base retriever, performance remains largely unchanged or slightly decreases, aligning with the lack of retrieval performance improvement.

\noindent
\textbf{Human-GPT-4 Evaluation Correlation}
To verify the soundness of the GPT-4-based evaluation in the aforementioned results, we conducted human annotation on a subset of the dataset using identical criteria and calculated the correlation with GPT-4-based scores. 
We randomly sampled 20 queries each from the Story and RAG-QA datasets.
We chose BM25 as the baseline search model and selected the number of passages that yielded the most significant performance boost for LightPAL. 
For each query, summaries generated by MegaBeam-Mistral-7B using BM25, BM25+LightPAL, and BM25+PromptRank were randomly shuffled and annotated by three natural language processing researchers.
Two annotators each evaluated 10 different queries per dataset, while the third annotator covered all queries, resulting in two annotations per query. 
We then calculated the correlation between the average of these scores and the GPT-4-based scores. The annotation instructions and interface are detailed in Appendix~\ref{appendix:human_annotation}. 
Table~\ref{tab:correlation} presents the correlation coefficients between human judgment and GPT-4 evaluation. 
Except for the RAG-QA and Div combination, we observe Spearman's $\rho$ values ranging from approximately 0.3 to 0.5 across the other conditions. These moderate correlations suggest that GPT-4-based evaluations demonstrate a reasonable alignment with human judgments and its soundness.

\begin{figure}[!t]
    \centering
    \includegraphics[width=7.0cm]{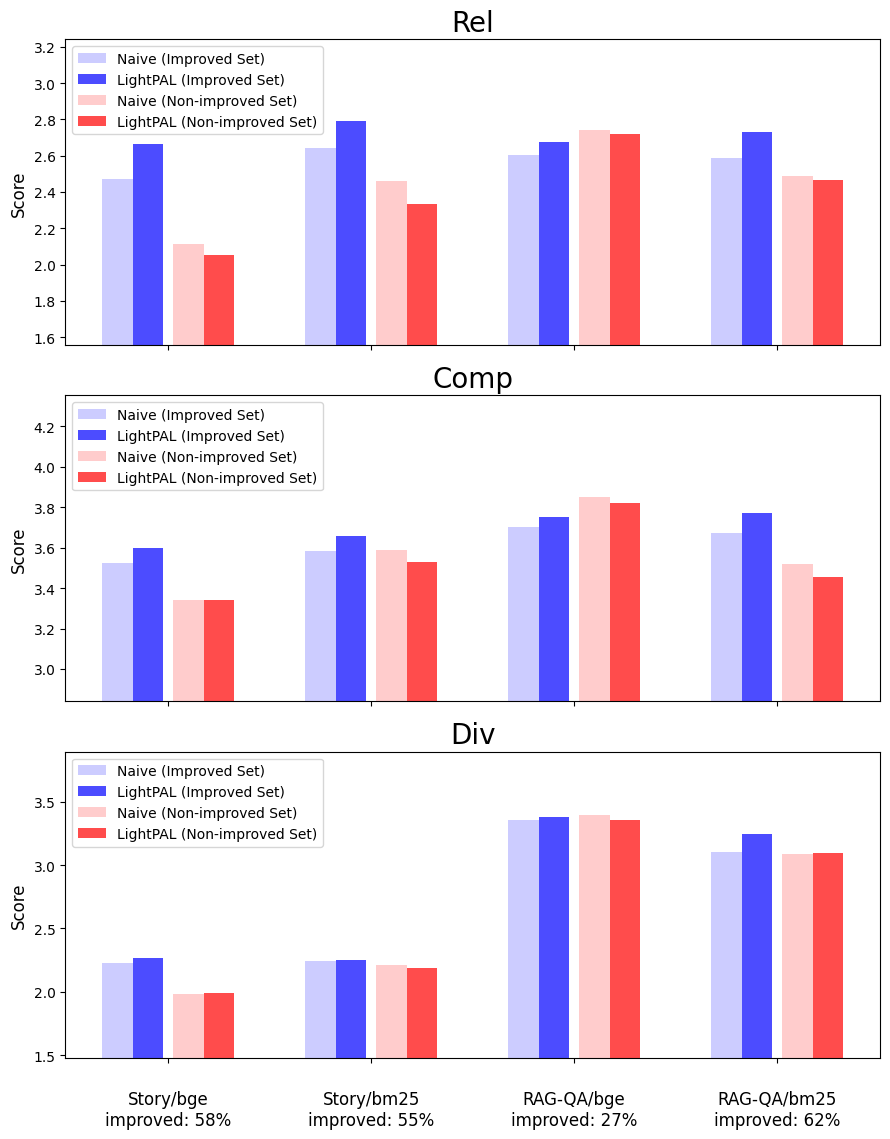}
    \caption{Summary quality plot for samples with improved retrieval by LightPAL (Improved Set) versus those without (Non-improved Set). Each group represents a combination of dataset and base retriever, with the percentage of samples showing improved retrieval performance noted below.}
    \label{fig:detailed_imp_vs_nonimp}
\end{figure}

\subsection{Detailed Analysis}
\label{sec:exp/detailed_analysis}
\noindent
\textbf{Does better retrieval performance really mean better summary quality?}
To improve the quality of summaries in ODMDS task, we proposed a method to enhance retrieval performance. But does better retrieval truly lead to higher quality summaries? To answer this question, we divided samples from Story and RAG-QA datasets into an Improved Set (where LightPAL improved Recall@$k$) and a Non-Improved Set. Figure~\ref{fig:detailed_imp_vs_nonimp} shows the results of our analysis. The Improved Set (shown in blue bars) demonstrates enhanced summary quality through LightPAL, particularly in terms of Relevance. Conversely, the Non-Improved Set shows either no change or a decline in performance. Notably, up to 62\% of samples in the RAG-QA dataset with BM25 retriever showed improved retrieval performance, indicating that a significant number of samples require iterative retrieval method like LightPAL and PromptRank.

\noindent
\textbf{Effective Cases in LightPAL Retrieval Performance.}
To analyze when LightPAL succeeds or fails across various queries, documents, and base retrieval models, we present two cases in Figure~\ref{fig:detailed_examples}. 
LightPAL enhances retrieval by 1) covering passages without distinctive query keywords and 2) selecting topically similar passages when multiple passages share common keywords. The latter is particularly beneficial compared to sparse retrievers like BM25. However, LightPAL may potentially degrade performance when distinctive keywords are present in both the target passage and the query, or when the system can accurately grasp the topic without assistance.
For examples of generated summaries, please refer to Appendix~\ref{appendix:gen_summs}.

\begin{figure}[t!]
    \centering
    \begin{tcolorbox}[
      enhanced,
      colback=white,
      colframe=black,
      boxrule=0.5pt,
      arc=0pt,
      outer arc=0pt,
      leftupper=2mm,
      rightupper=2mm,
      top=1mm,
      bottom=1mm,
      toptitle=0mm,
      bottomtitle=0mm,
      before upper={\parskip=0.3em},
      width=\linewidth
    ]
    \textbf{CASE 1: passages do not contain named entities in the query} \\
    \uline{Query}: What is the dynamic between Captain Midas and Mister Spinelli in the story of CAPTAIN MIDAS? \\
    \uline{Naive (bge-large-en-v1.5)}: \\
    \redcross: The \textcolor{red}{\textit{captain}}'s voice. Calm, brief. It sent a tremor... (a section of different story) \\ 
    \uline{LightPAL}: \\
    \greencheck: The first thing about the derelict...
    \end{tcolorbox} 

    \begin{tcolorbox}[
      enhanced,
      colback=white,
      colframe=black,
      boxrule=0.5pt,
      arc=0pt,
      outer arc=0pt,
      leftupper=2mm,
      rightupper=2mm,
      top=1mm,
      bottom=1mm,
      toptitle=0mm,
      bottomtitle=0mm,
      before upper={\parskip=0.3em},
      width=\linewidth
    ]
    \textbf{CASE 2: misunderstanding of the topic} \\
    \uline{Query:} How to see how much time you spend on a game on \textbf{xbox}? \\
    \uline{Naive (bm25):} \\
    \redcross: you have to get Nintendo's permission, buy from Nintendo's factories, .... If you want onto the \textcolor{red}{\textit{XBox}}, sure there's XBLA...\\
    \uline{LightPAL:} \\
    \greencheck: go to the friends tab and the choose...
    \end{tcolorbox}

\caption{Case study of retrieved passages using LightPAL and a naive base retriever.}
\label{fig:detailed_examples}
\end{figure}
\section{Conclusion}
In this paper, we proposed LightPAL, a lightweight passage retrieval method for Open-Domain Multi-Document Summarization. Our method constructs a graph representing passage relationships using language models during indexing. At retrieval time, random walk via Personalized PageRank on this graph enables low-latency retrieval of diverse relevant information without runtime LLM inference. Experiments on ODMDS and long-form QA benchmarks demonstrate that LightPAL outperforms naive sparse/dense retrievers in both retrieval and summarization metrics, while achieving higher efficiency compared to an iterative retrieval and reasoning approach for multi-hop QA.
\newpage
\section{Limitations}
\textbf{Variation of LLMs for summary generation. }
This paper used representative LLMs capable of handling long contexts (Meta-Llama-3.1-8B-Instruct and MegaBeam-Mistral-7B-512) for summary generation. A comprehensive evaluation using diverse language models as summarizers would be preferable to further validate the effectiveness of LightPAL and identify potential limitations.

\noindent
\textbf{Limited comparison with other iterative reasoning and retrieval approaches.}
PromptRank~\cite{related/mqa/KhalifaLLL023} is not the only approach that iteratively performs reasoning and retrieval. While PromptRank was chosen as a representative approach for comparison, other methods exist, such as those using Chain-of-Thoughts~\cite{related/mqa/TrivediBKS23} or LLMs as graph traversal agents~\cite{related/mqa/Wang}. A more comprehensive comparison with these alternative approaches could provide additional insights into the strengths and limitations of our proposed method.

\noindent
\textbf{Selective application of LightPAL. }
LightPAL does not universally enhance retrieval performance across all samples. Additionally, its effectiveness varies depending on the dataset and base retriever used. To maximize the utility of our proposed method, developing criteria for when to apply LightPAL would be beneficial. For example, if the similarity between the query and retrieved passages exceeds a predetermined threshold, we might conclude that the base retriever is adequate without applying LightPAL.
\section*{Acknowledgments}
We thank Takuya Tamura (NEC Corporation) for the discussion of this research.

\bibliography{myref}

\begin{thebibliography}{21}
\providecommand{\natexlab}[1]{#1}

\bibitem[{Das et~al.(2019)Das, Dhuliawala, Zaheer, and McCallum}]{related/mqa/DasDZM19}
Rajarshi Das, Shehzaad Dhuliawala, Manzil Zaheer, and Andrew McCallum. 2019.
\newblock \href {https://openreview.net/forum?id=HkfPSh05K7} {Multi-step retriever-reader interaction for scalable open-domain question answering}.
\newblock In \emph{7th International Conference on Learning Representations, {ICLR} 2019, New Orleans, LA, USA, May 6-9, 2019}. OpenReview.net.

\bibitem[{Edge et~al.(2024)Edge, Trinh, Cheng, Bradley, Chao, Mody, Truitt, and Larson}]{related/rag/graph_ret/GraphRAG}
Darren Edge, Ha~Trinh, Newman Cheng, Joshua Bradley, Alex Chao, Apurva Mody, Steven Truitt, and Jonathan Larson. 2024.
\newblock \href {https://doi.org/10.48550/ARXIV.2404.16130} {From local to global: {A} graph {RAG} approach to query-focused summarization}.
\newblock \emph{CoRR}, abs/2404.16130.

\bibitem[{Giorgi et~al.(2023)Giorgi, Soldaini, Wang, Bader, Lo, Wang, and Cohan}]{related/odmds/pioneer}
John~M. Giorgi, Luca Soldaini, Bo~Wang, Gary~D. Bader, Kyle Lo, Lucy~Lu Wang, and Arman Cohan. 2023.
\newblock \href {https://doi.org/10.18653/V1/2023.FINDINGS-EMNLP.549} {Open domain multi-document summarization: {A} comprehensive study of model brittleness under retrieval}.
\newblock In \emph{Findings of the Association for Computational Linguistics: {EMNLP} 2023, Singapore, December 6-10, 2023}, pages 8177--8199. Association for Computational Linguistics.

\bibitem[{Han et~al.(2024)Han, Zhang, Qi, Xu, Wang, Liu, Wang, Min, and Castelli}]{experiment/dataset/LFRQA}
Rujun Han, Yuhao Zhang, Peng Qi, Yumo Xu, Jenyuan Wang, Lan Liu, William~Yang Wang, Bonan Min, and Vittorio Castelli. 2024.
\newblock \href {https://arxiv.org/abs/2407.13998} {Rag-qa arena: Evaluating domain robustness for long-form retrieval augmented question answering}.
\newblock \emph{Preprint}, arXiv:2407.13998.

\bibitem[{Hsieh et~al.(2024)Hsieh, Sun, Kriman, Acharya, Rekesh, Jia, Zhang, and Ginsburg}]{related/misc/RULER}
Cheng{-}Ping Hsieh, Simeng Sun, Samuel Kriman, Shantanu Acharya, Dima Rekesh, Fei Jia, Yang Zhang, and Boris Ginsburg. 2024.
\newblock \href {https://doi.org/10.48550/ARXIV.2404.06654} {{RULER:} what's the real context size of your long-context language models?}
\newblock \emph{CoRR}, abs/2404.06654.

\bibitem[{Jeh and Widom(2003{\natexlab{a}})}]{method/orig_ppr}
Glen Jeh and Jennifer Widom. 2003{\natexlab{a}}.
\newblock \href {https://doi.org/10.1145/775152.775191} {Scaling personalized web search}.
\newblock In \emph{Proceedings of the Twelfth International World Wide Web Conference, {WWW} 2003, Budapest, Hungary, May 20-24, 2003}, pages 271--279. {ACM}.

\bibitem[{Jeh and Widom(2003{\natexlab{b}})}]{related/misc/PPR}
Glen Jeh and Jennifer Widom. 2003{\natexlab{b}}.
\newblock \href {https://doi.org/10.1145/775152.775191} {Scaling personalized web search}.
\newblock In \emph{Proceedings of the Twelfth International World Wide Web Conference, {WWW} 2003, Budapest, Hungary, May 20-24, 2003}, pages 271--279. {ACM}.

\bibitem[{Khalifa et~al.(2023)Khalifa, Logeswaran, Lee, Lee, and Wang}]{related/mqa/KhalifaLLL023}
Muhammad Khalifa, Lajanugen Logeswaran, Moontae Lee, Honglak Lee, and Lu~Wang. 2023.
\newblock \href {https://doi.org/10.18653/V1/2023.ACL-LONG.885} {Few-shot reranking for multi-hop {QA} via language model prompting}.
\newblock In \emph{Proceedings of the 61st Annual Meeting of the Association for Computational Linguistics (Volume 1: Long Papers), {ACL} 2023, Toronto, Canada, July 9-14, 2023}, pages 15882--15897. Association for Computational Linguistics.

\bibitem[{Li et~al.(2024)Li, He, Guo, Bu, Bai, Liu, Liu, Qu, Li, Ouyang, Su, and Zheng}]{related/rag/graph_trav/GraphReader}
Shilong Li, Yancheng He, Hangyu Guo, Xingyuan Bu, Ge~Bai, Jie Liu, Jiaheng Liu, Xingwei Qu, Yangguang Li, Wanli Ouyang, Wenbo Su, and Bo~Zheng. 2024.
\newblock \href {https://doi.org/10.48550/ARXIV.2406.14550} {Graphreader: Building graph-based agent to enhance long-context abilities of large language models}.
\newblock \emph{CoRR}, abs/2406.14550.

\bibitem[{Liu et~al.(2023)Liu, Iter, Xu, Wang, Xu, and Zhu}]{experiment/metric/geval}
Yang Liu, Dan Iter, Yichong Xu, Shuohang Wang, Ruochen Xu, and Chenguang Zhu. 2023.
\newblock \href {https://doi.org/10.18653/V1/2023.EMNLP-MAIN.153} {G-eval: {NLG} evaluation using gpt-4 with better human alignment}.
\newblock In \emph{Proceedings of the 2023 Conference on Empirical Methods in Natural Language Processing, {EMNLP} 2023, Singapore, December 6-10, 2023}, pages 2511--2522. Association for Computational Linguistics.

\bibitem[{Mavi et~al.(2022)Mavi, Jangra, and Jatowt}]{related/mqa/survey}
Vaibhav Mavi, Anubhav Jangra, and Adam Jatowt. 2022.
\newblock \href {https://doi.org/10.48550/ARXIV.2204.09140} {A survey on multi-hop question answering and generation}.
\newblock \emph{CoRR}, abs/2204.09140.

\bibitem[{Qi et~al.(2019)Qi, Lin, Mehr, Wang, and Manning}]{related/mqa/QiLMWM19}
Peng Qi, Xiaowen Lin, Leo Mehr, Zijian Wang, and Christopher~D. Manning. 2019.
\newblock \href {https://doi.org/10.18653/V1/D19-1261} {Answering complex open-domain questions through iterative query generation}.
\newblock In \emph{Proceedings of the 2019 Conference on Empirical Methods in Natural Language Processing and the 9th International Joint Conference on Natural Language Processing, {EMNLP-IJCNLP} 2019, Hong Kong, China, November 3-7, 2019}, pages 2590--2602. Association for Computational Linguistics.

\bibitem[{Santhanam et~al.(2022)Santhanam, Khattab, Saad-Falcon, Potts, and Zaharia}]{experiment/dataset/LoTTE}
Keshav Santhanam, Omar Khattab, Jon Saad-Falcon, Christopher Potts, and Matei Zaharia. 2022.
\newblock \href {https://doi.org/10.18653/v1/2022.naacl-main.272} {{C}ol{BERT}v2: Effective and efficient retrieval via lightweight late interaction}.
\newblock In \emph{Proceedings of the 2022 Conference of the North American Chapter of the Association for Computational Linguistics: Human Language Technologies}, pages 3715--3734, Seattle, United States. Association for Computational Linguistics.

\bibitem[{Sarthi et~al.(2024)Sarthi, Abdullah, Tuli, Khanna, Goldie, and Manning}]{related/rag/graph_ret/raptor}
Parth Sarthi, Salman Abdullah, Aditi Tuli, Shubh Khanna, Anna Goldie, and Christopher~D Manning. 2024.
\newblock \href {https://openreview.net/forum?id=GN921JHCRw} {{RAPTOR}: Recursive abstractive processing for tree-organized retrieval}.
\newblock In \emph{The Twelfth International Conference on Learning Representations}.

\bibitem[{Trivedi et~al.(2023)Trivedi, Balasubramanian, Khot, and Sabharwal}]{related/mqa/TrivediBKS23}
Harsh Trivedi, Niranjan Balasubramanian, Tushar Khot, and Ashish Sabharwal. 2023.
\newblock \href {https://doi.org/10.18653/V1/2023.ACL-LONG.557} {Interleaving retrieval with chain-of-thought reasoning for knowledge-intensive multi-step questions}.
\newblock In \emph{Proceedings of the 61st Annual Meeting of the Association for Computational Linguistics (Volume 1: Long Papers), {ACL} 2023, Toronto, Canada, July 9-14, 2023}, pages 10014--10037. Association for Computational Linguistics.

\bibitem[{Wang et~al.(2024{\natexlab{a}})Wang, Lipka, Rossi, Siu, Zhang, and Derr}]{related/rag/graph_trav/KGP}
Yu~Wang, Nedim Lipka, Ryan~A. Rossi, Alexa~F. Siu, Ruiyi Zhang, and Tyler Derr. 2024{\natexlab{a}}.
\newblock \href {https://doi.org/10.1609/AAAI.V38I17.29889} {Knowledge graph prompting for multi-document question answering}.
\newblock In \emph{Thirty-Eighth {AAAI} Conference on Artificial Intelligence, {AAAI} 2024, Thirty-Sixth Conference on Innovative Applications of Artificial Intelligence, {IAAI} 2024, Fourteenth Symposium on Educational Advances in Artificial Intelligence, {EAAI} 2014, February 20-27, 2024, Vancouver, Canada}, pages 19206--19214. {AAAI} Press.

\bibitem[{Wang et~al.(2024{\natexlab{b}})Wang, Lipka, Rossi, Siu, Zhang, and Derr}]{related/mqa/Wang}
Yu~Wang, Nedim Lipka, Ryan~A. Rossi, Alexa~F. Siu, Ruiyi Zhang, and Tyler Derr. 2024{\natexlab{b}}.
\newblock \href {https://doi.org/10.1609/AAAI.V38I17.29889} {Knowledge graph prompting for multi-document question answering}.
\newblock In \emph{Thirty-Eighth {AAAI} Conference on Artificial Intelligence, {AAAI} 2024, Thirty-Sixth Conference on Innovative Applications of Artificial Intelligence, {IAAI} 2024, Fourteenth Symposium on Educational Advances in Artificial Intelligence, {EAAI} 2014, February 20-27, 2024, Vancouver, Canada}, pages 19206--19214. {AAAI} Press.

\bibitem[{Yang et~al.(2024)Yang, Wang, Wei, Wang, and Wen}]{related/misc/PPR_algo}
Mingji Yang, Hanzhi Wang, Zhewei Wei, Sibo Wang, and Ji-Rong Wen. 2024.
\newblock \href {https://doi.org/10.1109/TKDE.2024.3376000} {Efficient algorithms for personalized pagerank computation: A survey}.
\newblock \emph{IEEE Transactions on Knowledge and Data Engineering}, pages 1--20.

\bibitem[{Yang et~al.(2018)Yang, Qi, Zhang, Bengio, Cohen, Salakhutdinov, and Manning}]{related/mqa/hotpotqa}
Zhilin Yang, Peng Qi, Saizheng Zhang, Yoshua Bengio, William~W. Cohen, Ruslan Salakhutdinov, and Christopher~D. Manning. 2018.
\newblock \href {https://doi.org/10.18653/V1/D18-1259} {Hotpotqa: {A} dataset for diverse, explainable multi-hop question answering}.
\newblock In \emph{Proceedings of the 2018 Conference on Empirical Methods in Natural Language Processing, Brussels, Belgium, October 31 - November 4, 2018}, pages 2369--2380. Association for Computational Linguistics.

\bibitem[{Zhou et~al.(2023)Zhou, Shi, Zhang, Liu, Zhao, and Cohan}]{related/odmds/bench}
Yijie Zhou, Kejian Shi, Wencai Zhang, Yixin Liu, Yilun Zhao, and Arman Cohan. 2023.
\newblock \href {https://doi.org/10.48550/ARXIV.2309.08960} {Odsum: New benchmarks for open domain multi-document summarization}.
\newblock \emph{CoRR}, abs/2309.08960.

\bibitem[{Zhu et~al.(2021)Zhu, Lei, Wang, Zheng, Poria, and Chua}]{related/qa/survey}
Fengbin Zhu, Wenqiang Lei, Chao Wang, Jianming Zheng, Soujanya Poria, and Tat{-}Seng Chua. 2021.
\newblock \href {https://arxiv.org/abs/2101.00774} {Retrieving and reading: {A} comprehensive survey on open-domain question answering}.
\newblock \emph{CoRR}, abs/2101.00774.

\end{thebibliography}
\clearpage
\appendix
\section{Prompts for Generation/Evalaution}
\label{appendix:prompts}
\subsection{Prompts for summary evaluation}
\label{appendix:prompts/geval}

\begin{lstlisting}[basicstyle=\ttfamily\scriptsize, breaklines=true, frame=single, caption={Prompt for Relevance Evaluation}, captionpos=b]]
You will be given separated chunks of text and a question. You will then be given one answer written for the question based on the text.

Your task is to rate the answer on one metric.

Please make sure you read and understand these instructions carefully. Please keep this document open while reviewing, and refer to it as needed.


Evaluation Criteria:

Relevance (1-5) - selection of important content from the source. The answer should include only important information regarding the question from the source text. Annotators were instructed to penalize answers which contained redundancies and excess information.

Evaluation Steps:

1. Read the answer and the source text carefully.
2. Compare the answer to the source document and identify the main points of the source text regarding the question.
3. Assess how well the answer covers the main points of the source text regarding the question, and how much irrelevant or redundant information it contains.
4. Assign a relevance score from 1 to 5.


Question:

{{Question}}

Source Text:

{{Document}}

Answer:

{{Summary}}


Evaluation Form (scores ONLY):

- Relevance:
\end{lstlisting}

\begin{lstlisting}[basicstyle=\ttfamily\scriptsize, breaklines=true, frame=single, caption={Prompt for Diversity Evaluation}, captionpos=b]]
You will be given a question and one answer written for the question.

Your task is to rate the answer on one metric.

Please make sure you read and understand these instructions carefully. Please keep this document open while reviewing, and refer to it as needed.


Evaluation Criteria:

Diversity (1-5) - How varied and insightful is the answer in providing different perspectives on the question? Assess whether the answer introduces multiple viewpoints and facilitates a broader understanding of the topic.

Evaluation Steps:

1. Read and Understand: Begin by reading the question and the answer to understand the overall approach.
2. Identify Perspectives: Determine whether the answer includes multiple viewpoints, reflecting different interpretations or aspects of the topic.
3. Evaluate Depth of Insights: Assess the richness of insights provided by these viewpoints, considering how they contribute to a deeper understanding of the topic.
4. Assign a Score: Based on your evaluation, assign a diversity score from 1 to 5, reflecting the variety and depth of perspectives and insights presented in the answer.


Question:

{{Question}}

Answer:

{{Summary}}


Evaluation Form (scores ONLY):

- Diversity:
\end{lstlisting}

\begin{lstlisting}[basicstyle=\ttfamily\scriptsize, breaklines=true, frame=single, caption={Prompt for Comprehensiveness Evaluation}, captionpos=b]]
You will be given a question and one answer written for the question.

Your task is to rate the answer on one metric.

Please make sure you read and understand these instructions carefully. Please keep this document open while reviewing, and refer to it as needed.


Evaluation Criteria:

Comprehensiveness (1-5) - Does the answer cover all aspect of the question? Evaluate whether the answer provides a thorough representation of information that directly addresses the question.

Evaluation Steps:

1. Read and Understand: Start by reading the question and the provided answer thoroughly to grasp the context and scope.
2. Verify Coverage: Check if the answer directly addresses all aspect of the question.
3. Assess Detail: Evaluate the detail in the explanations provided, ensuring that they are sufficient to fully address the question.
4. Assign a Score: Based on your evaluation, assign a comprehensiveness score from 1 to 5, reflecting how completely the answer addresses the question.


Question:

{{Question}}

Answer:

{{Summary}}


Evaluation Form (scores ONLY):

- Comprehensiveness:
\end{lstlisting}

\subsection{Prompts for summary generation}
\label{appendix:prompts/summary}
\begin{lstlisting}[basicstyle=\ttfamily\scriptsize, breaklines=true, frame=single, caption={Prompt for summary generation in Story dataset}, captionpos=b]]
You are a helpful assistant that gives long answer to question based on long stories. Write an answer based on the following question and the separated sections in the stories. Please write the answer in approximately {{n_summ_words}} words. When summarizing, use only information explicitly stated in the provided story. Do not use your own knowledge or make inferences beyond what is directly stated in the text.

# STORY: {{story_idx}}
## SECTION: {passage_idx}}
{{passage}}
... (repeated for all retrieved passages)

QUESTION: {{query}}
SUMMARY:
\end{lstlisting}

\begin{lstlisting}[basicstyle=\ttfamily\scriptsize, breaklines=true, frame=single, caption={Prompt for summary generation in Meeting dataset}, captionpos=b]]
You are a helpful assistant that gives long answer to question based on long meetings. Write an answer based on the following question and the separated sections in the meetings. Please write the answer in approximately {{n_summ_words}} words. When summarizing, use only information explicitly stated in the provided meetings. Do not use your own knowledge or make inferences beyond what is directly stated in the text.

# MEETING: {{meeting_idx}}
## SECTION: {passage_idx}}
{{passage}}
... (repeated for all retrieved passages)

QUESTION: {{query}}
SUMMARY:
\end{lstlisting}

\begin{lstlisting}[basicstyle=\ttfamily\scriptsize, breaklines=true, frame=single, caption={Prompt for summary generation in RAG-QA dataset}, captionpos=b]]
You are a helpful assistant that gives long answer to question based on documents. Write an answer based on the following question and the documents. Please write the answer in approximately {{n_summ_words}} words. When summarizing, use only information explicitly stated in the provided documents. Do not use your own knowledge or make inferences beyond what is directly stated in the text.
    
# DOCUMENT: {{passage_idx}}
{{passage}}
... (repeated for all retrieved passages)

QUESTION: {{query}}
SUMMARY:
\end{lstlisting}

\section{Human Annotation Procedure}
\label{appendix:human_annotation}
\begin{figure*}[!t]
    \centering
    \includegraphics[width=16cm]{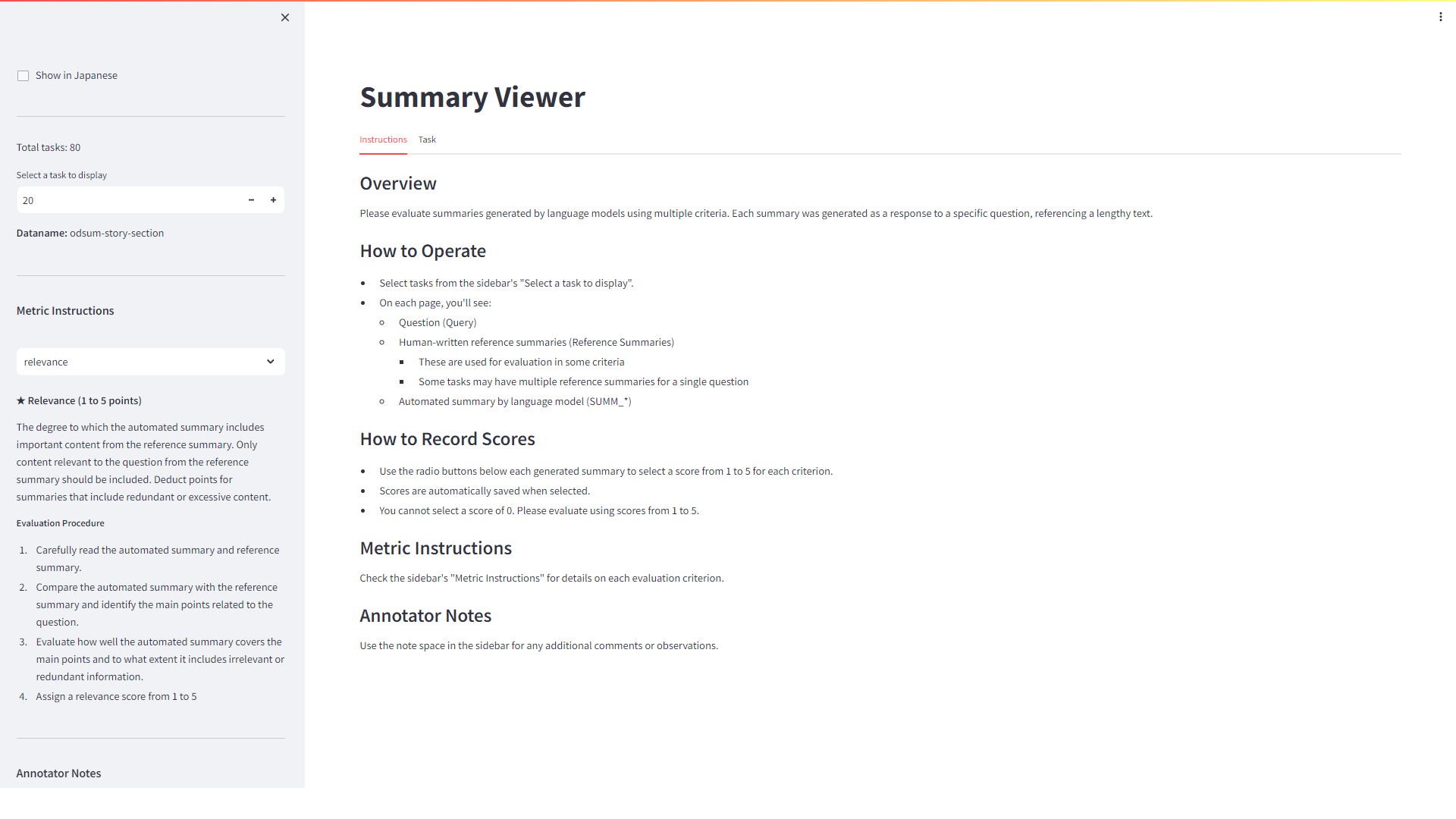}
    \caption{Annotation instruction}
    \label{fig:human_anno_inst}
\end{figure*}

\begin{figure*}[!t]
    \centering
    \includegraphics[width=16cm]{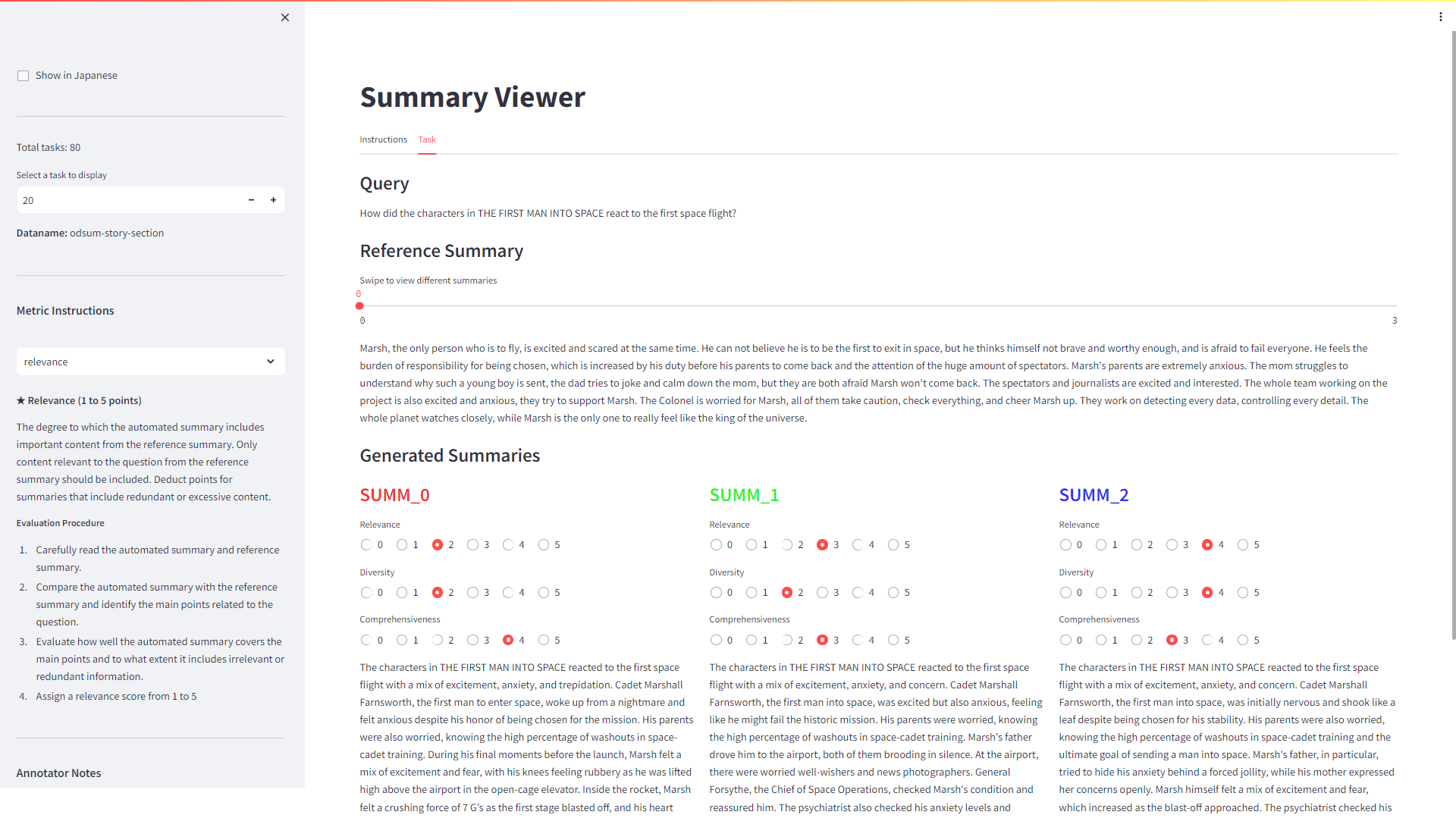}
    \caption{Annotation interface}
    \label{fig:human_anno_task}
\end{figure*}

Figure \ref{fig:human_anno_inst} and Figure \ref{fig:human_anno_task} show the instructions for annotation and the annotation interface, respectively. The human annotation was conducted using exactly the same aspects (relevance, comprehensiveness, diversity) as the GPT-4 evaluation, and the descriptions for each aspect were identical to those used in the prompts for GPT-4.

\section{Examples of Generated Summaries}
\label{appendix:gen_summs}
Figures~\ref{fig:summ_example_ragqa_better} and \ref{fig:summ_example_story_better} show examples of generated summaries when LightPAL is applied to the base retriever (BM25/bge-large-en-v1.5).

\begin{figure*}[t!]
    \centering
    \begin{tcolorbox}[
      enhanced,
      colback=white,
      colframe=black,
      boxrule=0.5pt,
      arc=0pt,
      outer arc=0pt,
      leftupper=2mm,
      rightupper=2mm,
      top=1mm,
      bottom=1mm,
      toptitle=0mm,
      bottomtitle=0mm,
      before upper={\parskip=0.3em},
      width=\linewidth
    ]
    \uline{\textbf{Dataset}}: RAG-QA \\[0.5em]
    \uline{\textbf{Query}}: what baseball rules or calls have become customary despite them not being in the rule book? \\
    \uline{\textbf{Reference Summary}}: In baseball, some of the unwritten rules include not stealing bases when there is a large lead, not placing your bat on the plate following a walk, and avoiding crossing over the pitching mound on your way back to the dugout after being called out. 
Other unspoken rules include that one should not walk a right-handed hitter to pitch to a left-handed hitter when a right-handed pitcher is on the mound. 
It is also considered taboo to talk about a no-hitter while it's in progress.  
Additionally, batters are encouraged to hit the ball where it is pitched.  
In a specific play referred to as "The neighborhood play" or sometimes "The Phantom double" it's accepted when a fielder does not actually make contact with the base during a double play attempt.  
Contradicting official MLB laws is the unwritten rule that a tie goes to the runner, even though the official rulebook indicates scenarios where either the ball must arrive before the batter or vice versa. 
Lastly, it's frowned upon to bunt when already holding a significant lead, as this is seen as an unnecessary strategy. \\
    \uline{\textbf{BM25}}: \\
    \colorbox{lightgray!90}{Scores: pre@k: 0.4000, rec@k: 0.3333, Rel: 2.3868, Comp: 3.3359, Div: 3.0131, ROUGE: 20.2281} \\
    \redcross: The concept of the "tie goes to the runner" in baseball is an example of an unwritten rule that contradicts official MLB rules. According to Rule 6.05j, a batter is out if they are tagged before reaching first base after a third strike or hitting a fair ball. However, Rule 7.08e states that any runner is out if they fail to reach the next base before being tagged. This results in a discrepancy where the outcome depends on whether the ball or the runner reaches the base first. Additionally, other unwritten rules include not walking a right-handed batter to pitch to a left-handed batter, not mentioning a no-hitter while it's in progress, maintaining a manager's detachment from players, and retaliating if a player is hit by a pitch. These rules, while not officially written, are widely accepted and practiced within the baseball community. \\ 
    \uline{\textbf{BM25 + LightPAL}}: \\
    \colorbox{lightgray!90}{Scores: pre@k: 0.8000, rec@k: 0.6667, Rel: 2.8995, Comp: 4.0558, Div: 3.9707, ROUGE: 26.4306} \\
    \greencheck: The concept of the "tie goes to the runner" is an example of a baseball rule that is often contradictory and not explicitly stated in the official MLB rules. According to Rule 6.05j, a batter is out if they are tagged before reaching first base after a third strike or hitting a fair ball. However, Rule 7.08e states that any runner is out if they fail to reach the next base before being tagged. This creates a situation where the outcome depends on whether the ball or the runner reaches the base first, effectively giving the tie to whichever team is closer to achieving their objective first. Additionally, there are several unwritten rules in baseball that have become customary despite not being officially written in the rulebook. These include not stealing when there is a large lead, not placing the bat on the plate after a walk, and not running over the pitching mound after an out. Other examples include the "neighborhood play" at second base, where the fielder doesn't touch the base during a double play attempt, and batters not making an effort to get out of the way of a pitch. An outfielder using the wall for support while attempting to catch a home run is also considered against the rules. These unwritten rules often revolve around sportsmanship and etiquette within the game.
    \end{tcolorbox}
    \caption{Examples of generated summaries in RAG-QA data using BM25 and BM25+LightPAL}
    \label{fig:summ_example_ragqa_better}
\end{figure*}

\begin{figure*}[t!]
    \centering
    \begin{tcolorbox}[
      enhanced,
      colback=white,
      colframe=black,
      boxrule=0.5pt,
      arc=0pt,
      outer arc=0pt,
      leftupper=2mm,
      rightupper=2mm,
      top=1mm,
      bottom=1mm,
      toptitle=0mm,
      bottomtitle=0mm,
      before upper={\parskip=0.3em},
      width=\linewidth
    ]
    \uline{\textbf{Dataset}}: Story \\[0.5em]
    \uline{\textbf{Query}}: What is the dynamic between Captain Midas and Mister Spinelli in the story of CAPTAIN MIDAS? \\
    \uline{\textbf{Reference Summary}}: Mister Spinelli is Third Officer under the command of Captain Midas and was the first to report the derelict ship and observe its potential to be claimed by the Maid. Spinelli is the first and only crew member to identify the metal from the abandoned ship as gold when he saw Captain Midas with it. The tension between Midas and Spinelli escalates and their relationship becomes antagonistic as both of them desire to benefit the most from this valuable gold and with Midas constantly pulling his authority over Spinelli. After Midas barrs him from being a part of the investigative crew, suspicion arises between the two as Spinelli suspects Midas wishes to keep the pot of gold for himself and Midas thinks that Spinelli may be telling others. This tension further escalates as Midas sees Spinelli nearly hit the trigger of the gun and in rage, the two end up fighting each other before Midas aimed a kick at his temple and killed him. 
...
However, this relationship later sours once Spinelli finds out that the discovered ship is made out of gold. The captain is wary because Spinelli is big and dangerous, preferring to keep an eye on him in case anything goes wrong. Spinelli, on the other hand, is suspicious of the captain and goes off to tell Zaleski even though the captain said to not mention the gold to any of the rest of the crew. The captain threatens to clap Spinelli to irons, but this threat does not work because the Third Officer chooses to disobey his orders anyways. Later, Spinelli and the captain get into conflict again, with Spinelli accusing the captain that he was planning to keep all of the treasure for himself. Spinelli assumes that everybody is faking it to keep the ship, but the captain knocks him away. The two of them fight brutally, and Captain Midas has to end up killing Spinelli.  \\
    \uline{\textbf{bge-large-en-v1.5}}: \\
    \colorbox{lightgray!90}{Scores: pre@k: 0.4000, rec@k: 0.5000, Rel: 2.7597, Comp: 3.5871, Div: 2.1482, ROUGE: 21.6848} \\
    \redcross: In the story of CAPTAIN MIDAS, the dynamic between Captain Midas and Mister Spinelli is one of suspicion and distrust. Captain Midas becomes aware of Spinelli's intense interest in the gold found on the derelict spaceship and suspects that Spinelli may be planning to steal it for himself. To prevent this, Captain Midas decides to keep the gold a secret from the rest of the crew and assigns Spinelli to remain on the ship with him, rather than allowing him to join the prize crew that will be boarding the derelict. However, Captain Midas also notes that Spinelli seems to be watching Zaleski, suggesting that Spinelli may be trying to prevent Zaleski from double-crossing him. Despite this, Captain Midas remains wary of Spinelli and keeps a close eye on him, ultimately deciding to relieve him of his duties and assign him to stay on the ship with him. \\
    \uline{\textbf{bge-large-en-v1.5 + LightPAL}}: \\
    \colorbox{lightgray!90}{Scores: pre@k: 0.8000, rec@k: 1.0000, Rel: 3.0666, Comp: 4.5831, Div: 2.8078, ROUGE: 20.0084} \\
    \greencheck: The dynamic between Captain Midas and Mister Spinelli in the story of CAPTAIN MIDAS is one of tension and mistrust. Initially, Captain Midas assigns Spinelli to stay on board the Maid with him, while Cohn and Zaleski take charge of the prize crew to bring the derelict ship into Callisto for salvage. However, Spinelli becomes suspicious that Captain Midas is planning to cheat him out of his rightful share of the treasure, and he begins to conspire with Zaleski to keep the treasure ship for themselves.

As the story progresses, Spinelli's actions become increasingly erratic and aggressive, and he eventually tries to double-cross Captain Midas by attempting to take control of the derelict ship. Captain Midas discovers Spinelli's plan and confronts him, leading to a violent altercation in which Captain Midas kills Spinelli.

Throughout the story, it is clear that Captain Midas and Spinelli have a complex and troubled relationship, marked by mutual distrust and a deep-seated desire for the treasure. Their dynamic is driven by the lure of wealth and the consequences of their actions, ultimately leading to tragic consequences for both characters.
    \end{tcolorbox}
    \caption{Examples of generated summaries in Story data using bge-large-en-v1.5 and bge-large-en-v1.5+LightPAL}
    \label{fig:summ_example_story_better}
\end{figure*}

\end{document}